# A Spectral Algorithm for Latent Junction Trees

**Ankur P. Parikh**
Carnegie Mellon University
apparikh@cs.cmu.edu

**Le Song**
Georgia Tech
lsong@cc.gatech.edu

**Mariya Ishteva**
Georgia Tech
mariya.ishteva@cc.gatech.edu

**Gabi Teodoru**
Gatsby Unit, UCL
gabi.teodoru@gmail.com

**Eric P. Xing**
Carnegie Mellon University
epxing@cs.cmu.edu

## Abstract

Latent variable models are an elegant framework for capturing rich probabilistic dependencies in many applications. However, current approaches typically parametrize these models using conditional probability tables, and learning relies predominantly on local search heuristics such as Expectation Maximization. Using tensor algebra, we propose an alternative parameterization of latent variable models (where the model structures are junction trees) that still allows for computation of marginals among observed variables. While this novel representation leads to a moderate increase in the number of parameters for junction trees of low treewidth, it lets us design a local-minimum-free algorithm for learning this parameterization. The main computation of the algorithm involves only tensor operations and SVDs which can be orders of magnitude faster than EM algorithms for large datasets. To our knowledge, this is the first provably consistent parameter learning technique for a large class of low-treewidth latent graphical models beyond trees. We demonstrate the advantages of our method on synthetic and real datasets.

## 1 Introduction

Latent variable models such as Hidden Markov Models (HMMs) (Rabiner & Juang, 1986), and Latent Dirichlet Allocation (Blei et al., 2003) have become a popular framework for modeling complex dependencies among variables. A latent variable can represent an abstract concept (such as topic or state), thus enriching the dependency structure among the observed variables while simultaneously allowing for a more tractable representation. Typically, a latent variable model is parameterized by a set of conditional probability tables (CPTs) each associated with an edge in the latent graph structure. For instance, an HMM can be parametrized compactly by a transition probability table and an observation probability table. By summing out the latent variables in the HMM, we obtain a fully connected graphical model for the observed variables.

Although the parametrization of latent variable models using CPTs is very compact, parameters in this representation can be difficult to learn. Compared to parameter learning in fully observed models which is either of closed form or convex (Koller & Friedman, 2009), most parameter learning algorithms for latent variable models resort to maximizing a non-convex objective via Expectation Maximization (EM) (Dempster et al., 1977). EM can get trapped in local optima and has slow convergence.

While EM explicitly learns the CPTs of a latent variable model, in many cases the goal of the model is primarily for prediction and thus the actual latent parameters are not needed. One example is determining splicing sites in DNA sequences (Asuncion & Newman, 2007). One can build a different latent variable model, such as an HMM, for each type of splice site from training data. A new sequence is then classified by determining which model it is most likely to have been generated by. Other examples include supervised topic modelling such as (Blei & McAuliffe, 2007; Lacoste-Julien et al., 2008; Zhu et al., 2009) and collaborative filtering (Su & Khoshgoftaar, 2009).

In these cases, it is natural to ask whether there exists an alternative representation/parameterization of a latent variable model where parameter learning can be done consistently and the representation remains tractable for inference among the observed variables. This question has been tackled recently by Hsu et al. (2009), Balle et al. (2011), and Parikh et al. (2011) who proposed spectral algorithms for local-minimum-free learning of HMMs, finite state transducers, and latent tree graphical models respectively. Unlike traditional parameter learning algorithms such as EM, spectral algorithms do not directly learn the CPTs of a latent variable model. Instead they learn an alternative parameterization (called the *observable representation*) which generally contains a larger number of parameters than the CPTs, but where computing observed marginals is still tractable. Moreover, these alternative parameters have the advantage that they only depend on observed variables and can therefore be directly estimated from data. Thus, parameter learning in the alternative representation is fast, local-minimum-free, and provably consistent. Furthermore, spectral algorithms can be generalized to nonparametric latent models (Song et al., 2010, 2011) where it is difficult to run EM.

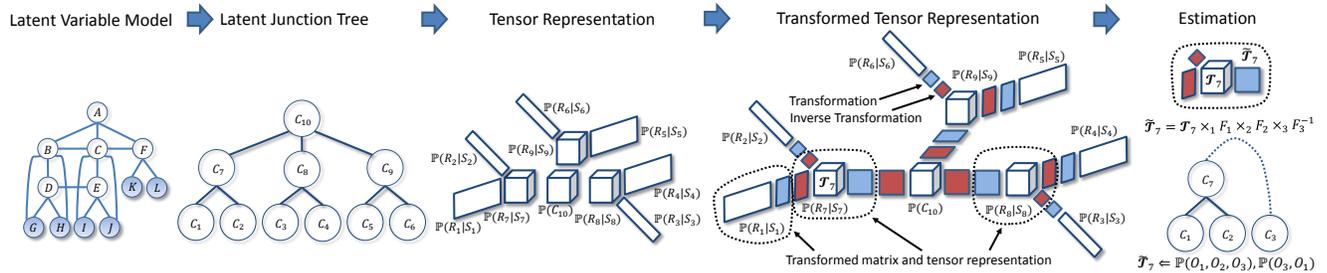

Figure 1: Our algorithm for local-minimum-free learning of latent variable models consist of four major steps. (1) First, we transform a model into a junction tree, such that each node in the junction tree corresponds to a maximal clique of variables in the triangulated graph of the original model. (2) Then we embed the clique potentials of the junction tree into higher order tensors and express the marginal distribution of the observed variables as a tensor-tensor/matrix multiplication according to the message passing algorithm. (3) Next we transform the tensor representation by inserting a pair of transformations between those tensor-tensor/matrix operations. Each pair of transformations is chosen so that they are inversions of each other. (4) Lastly, we show that each transformed representation is a function of only observed variables. Thus, we can estimate each individual transformed tensor quantity using samples from observed variables.

However, existing spectral algorithms apply only to restricted latent structures (HMMs and latent trees), while latent structures beyond trees, such as higher order HMMs (Kundu et al., 1989), factorial HMMs (Ghahramani & Jordan, 1997) and Dynamic Bayesian Networks (Murphy, 2002), are needed and have been proven useful in many real world problems. The challenges for generalizing spectral algorithms to general latent structured models include the larger factors, more complicated conditional independence structures, and the need to sum out multiple variables simultaneously.

The goal of this paper is to develop a new representation for latent variable models with structures beyond trees, and design a spectral algorithm for learning this representation. We will focus on latent junction trees; thus the algorithm is suitable for both directed and undirected models which can be transformed into junction trees. Concurrently to our work, Cohen et al. (2012) proposed a spectral algorithm for Latent Probabilistic Context Free Grammars (PCFGs). Latent PCFGs are not trees, but have many tree-like properties, and so the representation Cohen et al. (2012) propose does not easily extend to other non-tree models such as higher order/factorial HMMs that we consider here. Our more general approach requires more complex tensor operations, such as multi-mode inversion, that are not used in the latent PCFG case.

The key idea of our approach is to embed the clique potentials of the junction tree into higher order tensors such that the computation of the marginal probability of observed variables can be carried out via tensor operations. While this novel representation leads only to a moderate increase in the number parameters for junction trees of low treewidth, it allows us to design an algorithm that can recover a transformed version of the tensor parameterization and ensure that the joint probability of observed variables are computed correctly and consistently. The main computation of the algorithm involves only tensor operations and singular value decompositions (hence the name "spectral") which can be orders of magnitude faster than EM algorithms in large datasets. To our knowledge, this is the first provably consistent parameter learning technique for a large class of low-treewidth latent graphical models beyond trees. In our experiments with large scale synthetic datasets, we show that our spectral algorithm can be almost 2 orders of magnitude faster than EM while at the same achieving considerably better accuracy. Our spectral algorithm also achieves comparable accuracy to EM on real data.

**Organization of paper.** A high level overview of our approach is given in Figure 1. We first provide some background on tensor algebra and latent junction trees. We then derive the spectral algorithm by representing junction tree message passing with tensor operations, and then transform this representation into one that only depends on observed variables. Finally, we analyze the sample complexity of our method and evaluate it on synthetic and real datasets.

## 2 Tensor Notation

We first give an introduction to the tensor notation tailored to this paper. An $N$th order tensor is a multiway array with $N$ "modes", *i.e.*, $N$ indices

$\{i_1, i_2, \ldots, i_N\}$ are needed to access its entries. Subarrays of a tensor are formed when a subset of the indices is fixed, and we use a colon to denote all elements of a mode. For instance, $\mathcal{A}(i_1, \ldots, i_{n-1}, :, i_{n+1}, \ldots, i_N)$ are all elements in the $n$th mode of a tensor $\mathcal{A}$ with indices from the other $N-1$ modes fixed to $\{i_1, \ldots, i_{n-1}, i_{n+1}, \ldots, i_N\}$ respectively. Furthermore, we also use the shorthand $\boldsymbol{i}_{p:q} = \{i_p, i_{p+1}, \ldots, i_{q-1}, i_q\}$ for consecutive indices, e.g., $\mathcal{A}(i_1, \ldots, i_{n-1}, :, i_{n+1}, \ldots, i_N) = \mathcal{A}(\boldsymbol{i}_{1:n-1}, :, \boldsymbol{i}_{n+1:N})$.

**Labeling tensor modes with variables.** In contrast to the conventional tensor notation such as the one described in Kolda & Bader (2009), the ordering of the modes of a tensors will not be essential in this paper. We will use random variables to *label* the modes of a tensor: each mode will correspond to a random variable and what is important is to keep track of this correspondence. Therefore, we think two tensors are equivalent if they have the same set of labels and they can be obtained from each other by a permutation of the modes for which the labels are aligned.

In the matrix case this translates to $\boldsymbol{A}$ and $\boldsymbol{A}^\top$ being equivalent in the sense that $\boldsymbol{A}^\top$ carries the same information as $\boldsymbol{A}$, as long as we remember that the rows of $\boldsymbol{A}^\top$ are the columns of $\boldsymbol{A}$ and vice versa. We will use the following notation to denote this equivalence

$$\boldsymbol{A} \cong \boldsymbol{A}^\top \qquad (1)$$

Under this notation, the dimension (or the size) of a mode labeled by variable $X$ will be the same as the number of possible values for variable $X$. Furthermore, when we multiply two tensors together, we will always carry out the operation along (a set of) modes with matching labels.

**Tensor multiplication with mode labels.** Let $\mathcal{A} \in \mathbb{R}^{I_1 \times I_2 \times \cdots \times I_N}$ be an $N$th order tensor and $\mathcal{B} \in \mathbb{R}^{J_1 \times J_2 \times \cdots \times J_M}$ be an $M$th order tensor. If $X$ is a common mode label for both $\mathcal{A}$ and $\mathcal{B}$ (w.l.o.g. we assume that this is the first mode, implying also that $I_1 = J_1$), multiplying along this mode will give

$$\mathcal{C} = \mathcal{A} \times_X \mathcal{B} \in \mathbb{R}^{I_2 \times \cdots \times I_N \times J_2 \times \cdots \times J_M}, \qquad (2)$$

where the entries of $\mathcal{C}$ is defined as

$$\mathcal{C}(\boldsymbol{i}_{2:N}, \boldsymbol{j}_{2:M}) = \sum_{i=1}^{I_1} \mathcal{A}(i, \boldsymbol{i}_{2:N}) \mathcal{B}(i, \boldsymbol{j}_{2:M})$$

Similarly, we can multiply two tensors along multiple modes. Let $\boldsymbol{\sigma} = \{X_1, \ldots, X_k\}$ be an arbitrary set of $k$ modes ($k$ variables) shared by $\mathcal{A}$ and $\mathcal{B}$ (w.l.o.g. we assume these labels correspond to the first $k$ modes, and $I_1 = J_1, \ldots, I_k = J_k$ holds for the corresponding dimensions). Then multiplying $\mathcal{A}$ and $\mathcal{B}$ along $\boldsymbol{\sigma}$ results in

$$\mathcal{D} = \mathcal{A} \times_{\boldsymbol{\sigma}} \mathcal{B} \in \mathbb{R}^{I_{k+1} \times \cdots \times I_N \times J_{k+1} \times \cdots \times J_M}, \qquad (3)$$

where the entries of $\mathcal{D}$ are defined as
$$\mathcal{D}(\boldsymbol{i}_{k+1:N}, \boldsymbol{j}_{k+1:M}) = \sum_{\boldsymbol{i}_{1:k}} \mathcal{A}(\boldsymbol{i}_{1:k}, \boldsymbol{i}_{k+1:N}) \mathcal{B}(\boldsymbol{i}_{1:k}, \boldsymbol{j}_{k+1:M}).$$

Multi-mode multiplication can also be interpreted as reshaping the $\boldsymbol{\sigma}$ modes of $\mathcal{A}$ and $\mathcal{B}$ into a single mode and doing single-mode tensor multiplication. Furthermore, tensor multiplication with labels is symmetric in its arguments, i.e., $\mathcal{A} \times_{\boldsymbol{\sigma}} \mathcal{B} \cong \mathcal{B} \times_{\boldsymbol{\sigma}} \mathcal{A}$.

**Mode-specific identity tensor.** We now define our notion of identity tensor with respect to a set of modes $\boldsymbol{\sigma} = \{X_1, \ldots, X_K\}$. Let $\mathcal{A}$ be a tensor with mode labels containing $\boldsymbol{\sigma}$, and $\mathcal{I}_{\boldsymbol{\sigma}}$ be a tensor with $2K$ modes with mode labels $\{X_1, \ldots, X_K, X_1, \ldots, X_K\}$. Then $\mathcal{I}_{\boldsymbol{\sigma}}$ is an identity tensor with respect to modes $\boldsymbol{\sigma}$ if

$$\mathcal{A} \times_{\boldsymbol{\sigma}} \mathcal{I}_{\boldsymbol{\sigma}} \cong \mathcal{A}. \qquad (4)$$

One can also understand $\mathcal{I}_{\boldsymbol{\sigma}}$ using its matrix representation: flattening $\mathcal{I}_{\boldsymbol{\sigma}}$ with respect to $\boldsymbol{\sigma}$ (the first $\boldsymbol{\sigma}$ modes mapped to rows and the second $\boldsymbol{\sigma}$ modes mapped to columns) results in an identity matrix.

**Mode-specific tensor inversion.** Let $\mathcal{F}, \mathcal{F}^{-1} \in \mathbb{R}^{I_1 \times \cdots \times I_K \times I_{K+1} \times \cdots \times I_{K+K'}}$ be tensors of order $K + K'$, and both have two sets of mode labels $\boldsymbol{\sigma} = \{X_1, \ldots, X_K\}$ and $\boldsymbol{\omega}' = \{X_{K+1}, \ldots, X_{K+K'}\}$. Then $\mathcal{F}^{-1}$ is the inverse of $\mathcal{F}$ w.r.t. modes $\boldsymbol{\omega}$ if and only if

$$\mathcal{F} \times_{\boldsymbol{\omega}} \mathcal{F}^{-1} \cong \mathcal{I}_{\boldsymbol{\sigma}}. \qquad (5)$$

Multimode inversion can also be interpreted as reshaping $\mathcal{F}$ with respect to $\boldsymbol{\omega}$ into a matrix of size $(I_1 \ldots I_K) \times (I_{K+1} \ldots I_{K+K'})$, taking the inverse, and then rearranging back into a tensor. Thus the existence and uniqueness of this inverse can be characterized by the rank of the matricized version of $\mathcal{F}$.

**Mode-specific diagonal tensors.** We use $\boldsymbol{\delta}$ to denote an $N$-way relation: its entry $\boldsymbol{\delta}(\boldsymbol{i}_{1:N})$ at position $\boldsymbol{i}_{1:N}$ equals 1 when all indexes are the same ($i_1 = i_2 = \ldots = i_N$), and 0 otherwise. We will use $\oslash_d$ to denote repetition of an index $d$ times. For instance, we use $\mathbb{P}(\oslash_d X)$ to denote a $d$th order tensor where its entries at $(\boldsymbol{i}_{1:d})$th position are specified by $\boldsymbol{\delta}(\boldsymbol{i}_{1:d})\mathbb{P}(X = x_{i_1})$. A diagonal matrix with its diagonal equal to $\mathbb{P}(X)$ is then denoted as $\mathbb{P}(\oslash_2 X)$. Similarly, we can define a $(d + d')$th order tensor $\mathbb{P}(\oslash_d X | \oslash_{d'} Y)$ where its $(\boldsymbol{i}_{1:d}\boldsymbol{j}_{1:d'})$th entry corresponds to $\boldsymbol{\delta}(\boldsymbol{i}_{1:d})\boldsymbol{\delta}(\boldsymbol{j}_{1:d'})\mathbb{P}(X = x_{i_1}|Y = y_{j_1})$.

## 3 Latent Junction Trees

In this paper, we will focus on discrete latent variable models where the number of states, $k_h$, for each hidden variable is much smaller than the number of states, $k_o$, for each observed variable. Uppercase letters denote random variables (e.g., $X_i$) and lowercase letters their instantiations (e.g., $x_i$). A latent variable model defines a joint probability distribution over a set of variables $\mathscr{X} = \mathscr{O} \cup \mathscr{H}$. Here, $\mathscr{O}$ denotes the set of observed variables, $\{X_1, \ldots, X_{|\mathscr{O}|}\}$. $\mathscr{H}$ denotes the set of hidden variables, $\{X_{|\mathscr{O}|+1}, \ldots, X_{|\mathscr{H}|+|\mathscr{O}|}\}$.

We will focus on latent variable models where the structure of the model is a junction tree of low treewidth (Cowell et al., 1999). Each node $C_i$ in a junction tree corresponds to a subset (clique) of variables from the original graphical model. We will also use $C_i$ to denote the collection of variables contained in the node, i.e. $C_i \subset \mathscr{X}$. Let $\mathbb{C}$ denote the set of all clique nodes. The treewidth is then the size of a largest clique in a junction tree minus one, that is $t = \max_{C_i \in \mathbb{C}} |C_i| - 1$. Furthermore, we associate each edge in a junction tree with a separator set $S_{ij} := C_i \cap C_j$ which contains the common variables of the two cliques $C_i$ and $C_j$ it is connected to. If we condition on all variables in any $S_{ij}$, the variables on different sides of $S_{ij}$ will become independent.

Without loss of generality, we assume that each internal clique node in the junction tree has exactly 3 neighbors.[1] Then we can pick a clique $C_r$ as the root of the tree and reorient all edges away from the root to induce a topological ordering of the clique nodes. Given the ordering, the root node will have 3 children nodes, denoted as $C_{r_1}, C_{r_2}$ and $C_{r_3}$. Each other internal node $C_i$ will have a unique parent node, denoted as $C_{i_0}$, and 2 children nodes denoted as $C_{i_1}$ and $C_{i_2}$. Each leaf node $C_l$ is only connected with its unique parent node $C_{l_0}$. Furthermore, we can simplify the notation for the separator set between a node $C_i$ and its parent $C_{i_0}$ as $S_i = C_i \cap C_{i_0}$, omitting the index for the parent node. Then the remainder set of a node is defined as $R_i = C_i \setminus S_i$. We also assume w.l.o.g. that if $C_i$ is a leaf in the junction tree, $R_i$ consists of only observed variables. We will use $r_i$ to denote an instantiation of the set of variables in $R_i$. See Figure 2 for an illustration of notation.

Given a root and a topological ordering of the nodes in a junction tree, the joint distribution of all variables $\mathscr{X}$ can be factorized according to

$$\mathbb{P}(\mathscr{X}) = \prod_{i=1}^{|\mathscr{X}|} \mathbb{P}(R_i|S_i), \quad (6)$$

where each CPT $\mathbb{P}(R_i|S_i)$, also called a clique potential, corresponds to a node $C_i$. The number of parameters needed to specify the model is $O(|\mathbb{C}|k_o^t)$, linear in the number of cliques but exponential in the tree width $t$. Then the marginal distribution of the observed variables can be obtained by summing over the latent variables,

$$\mathbb{P}(\mathscr{O}) = \sum_{X_{|\mathscr{O}|+1}} \cdots \sum_{X_{|\mathscr{O}|+|\mathscr{H}|}} \left[\prod_{i=1}^{|\mathscr{X}|} \mathbb{P}(R_i|S_i)\right], \quad (7)$$

where we use $\sum_X \phi(X)$ to denote summation over all possible instantiations of $\phi(x)$ w.r.t. variable $X$. Note that each (non-leaf) remainder set $R_i$ contains a small subset of all latent variables. The presence of latent variables introduces complicated dependency between observed variables, while at the same time only a small number of parameters corresponding to the entries in the CPTs are needed to specify the model.

---
[1] If this is not the case, the derivation is similar but notationally much heavier.

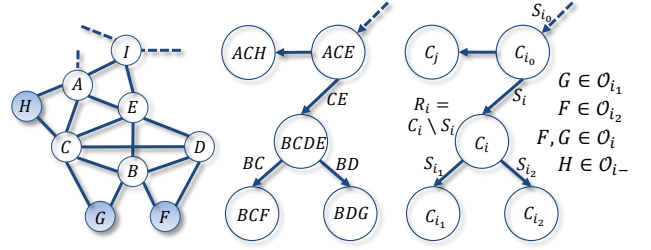

Figure 2: Example latent variable models with variables $\mathscr{X} = \{A, B, C, D, E, F, G, H, I, \ldots\}$, the observed variables are $\mathscr{O} = \{F, G, H, \ldots\}$ (only partially drawn). Its corresponding junction tree is shown in the middle panel. Corresponding to this junction tree, we also show the general notation for it in the rightmost panel.

The process of eliminating the latent variables in (7) can be carried out efficiently via message passing. More specifically, the summation can be broken up into local computation for each node in the junction tree. Each node only needs to sum out a small number of variables and then the intermediate result, called the message, is passed to its parent for further processing. In the end the root node incorporates all messages from its children and produces the final result $P(\mathscr{O})$. The local summation step, called the message update, can be generically written as[2]

$$\mathcal{M}(S_i) = \sum_{R_i} \mathbb{P}(R_i|S_i)\mathcal{M}(S_{i_1})\mathcal{M}(S_{i_2}) \quad (8)$$

where we use $\mathcal{M}(S_i)$ to denote the intermediate results of eliminating variables in the remainder set $R_i$. This message update is then carried out recursively according the reverse topological order of the junction tree until we reach the root node. The local summation step for the leaf nodes and root node can be viewed as special cases of (8). For a leaf node $C_l$, there is no incoming message from children nodes, and hence $\mathcal{M}(S_l) = \mathbb{P}(r_l|S_l)$; for the root node $C_r$, $S_r = \emptyset$ and $R_r = C_r$, and hence $\mathbb{P}(\mathscr{O}) = \mathcal{M}(\emptyset) = \sum_{\forall X \in C_r} \mathbb{P}(C_r)\mathcal{M}(S_{r_1})\mathcal{M}(S_{r_2})\mathcal{M}(S_{r_3})$.

**Example.** The message update at the internal node $C_{BCDE}$ in Figure 2 is
$$\mathcal{M}(\{C, E\}) = \sum_{B,D} \mathbb{P}(B, D|C, E)\mathbb{P}(f|B, C)\mathbb{P}(g|B, D).$$

## 4 Tensor Representation for Message Passing

Although the parametrization of latent junction trees using CPTs is very compact and inference (message passing) can be carried out efficiently, parameters in this representation can be difficult to learn. Since the likelihood of the observed data is no longer convex in the latent parameters, local search heuristics, such as EM, are often employed to learning the parameters. Therefore, our goal is to design a new representation for latent junction trees, such that subsequent learning can be carried out in a local-minimum-free fashion.

---
[2] For simplicity of notation, assume $C_i = S_i \cup S_{i_1} \cup S_{i_2}$.

In this section, we will develop a new representation for the message update in (8) by embedding each CPT $\mathbb{P}(R_i|S_i)$ into a higher order tensor $\mathcal{P}(C_i)$. As we will see, there will be two advantages to the tensor form. The first is that tensor multiplication can be used to compactly express the sum and product steps involved in message passing. As a very simplistic example, let $\boldsymbol{P}_{A|B} = \mathbb{P}(A|B)$ be a conditional probability matrix and $\boldsymbol{P}_B = \mathbb{P}(B)$ be a marginal probability vector. Then matrix-vector multiplication, $\boldsymbol{P}_{A|B}\boldsymbol{P}_B = P(A)$, sums out variable $B$. However, if we put the marginal probability of $B$ on the diagonal of a matrix, then $B$ will not be summed out: e.g., if $\boldsymbol{P}_{\oslash_2 B} = \mathbb{P}(\oslash_2 B)$, then $\boldsymbol{P}_{A|B}\boldsymbol{P}_{\oslash_2 B} = \mathbb{P}(A,B)$ (but now $B$ is no longer on the diagonal). We will leverage these facts to derive our tensor representation for message passing.

Moreover, we can then utilize tensor inversion to construct an alternate parameterization. In the very simplistic (matrix) example, note that $\mathbb{P}(A,B) = \boldsymbol{P}_{A|B}\boldsymbol{P}_{\oslash_2 B} = \boldsymbol{P}_{A|B}\boldsymbol{F}\boldsymbol{F}^{-1}\boldsymbol{P}_{\oslash_2 B}$. The invertible transformations $\boldsymbol{F}$ will give us an extra degree of freedom to allow us to design an alternate parameterization of the latent junction tree that is only a function of observed variables. This would not be possible in the traditional representation (Eq. 8).

### 4.1 Embed CPTs to higher order tensors

As we can see from (6), the joint probability distribution of all variables can be represented by a set of conditional distributions over just subsets of variables. Each one of this conditionals is a low order tensor. For example in Figure 2, the CPT corresponding to the clique node $C_{BCDE}$ would be a 4th order tensor $\mathbb{P}(B,D|C,E)$ where each variable corresponds to a different mode of the tensor. However, this representation is not suitable for deriving the observable representation since message passing cannot be defined easily using the tensor multiplication/sum connection shown above. Instead we will embed these tensors into even higher order tensors to facilitate the computation. The key idea is to introduce duplicate indexes using the mode-specific identity tensors, such that the sum-product steps in message updates can be expressed as tensor multiplications.

More specifically, the number of times a mode of the tensor is duplicated will depend on how many times the corresponding variable in the clique $C_i$ appears in the separator sets incident to $C_i$. We can define the count for a variable $X_j \in C_i$ as

$$d_{j,i} = \mathbb{I}[X_j \in S_i] + \mathbb{I}[X_j \in S_{i_1}] + \mathbb{I}[X_j \in S_{i_2}], \quad (9)$$

where $\mathbb{I}[\cdot]$ is an indicator function taking value 1 if its argument is true and 0 otherwise. Then the tensor representation of the node $C_i$ is

$$\mathcal{P}(C_i) := \\ \mathbb{P}(\underbrace{\ldots,(\oslash_{d_{j,i}}X_j),\ldots}_{\forall X_j \in R_i}|\underbrace{\ldots,(\oslash_{d_{j',i}}X_{j'}),\ldots}_{\forall X_{j'} \in S_i}), \quad (10)$$

where the labels for the modes of the tensor are the combined labels of the separator sets, i.e., $\{S_i, S_{i_1}, S_{i_2}\}$. The number of times a variable is repeated in the label set is exactly equal to $d_{j,i}$.

Essentially, tensor $\mathcal{P}(C_i)$ contains exactly the same information as the original CPT $\mathbb{P}(R_i|S_i)$. Furthermore, $\mathcal{P}(C_i)$ has a lot of zero entries, and the entries from $\mathbb{P}(R_i|S_i)$ are simply embedded in the higher order tensor $\mathcal{P}(C_i)$. Suppose all variables in node $C_i$ are latent variables each taking $k_h$ values. Then the number of entries needed to specify $\mathbb{P}(R_i|S_i)$ is $k_h^{|C_i|}$, while the high order tensor $\mathcal{P}(C_i)$ has $k_h^{d_i}$ entries where $d_i := \sum_{j:X_j \in C_i} d_{j,i}$ which is never smaller than $k_h^{|C_i|}$. In a sense, the parametrization using higher order tensor $\mathcal{P}(C_i)$ is less compact than the parametrization using the original CPTs. However, constructing the tensor $\mathcal{P}$ this way allows us to express the junction tree message update step in (8) as tensor multiplications (more details in the next section), and then we can leverage tools from tensor analysis to design a local-minimum-free learning algorithm.

The tensor representation for the leaf nodes and the root node are special cases of the representation in (10). The tensor representation at a leaf node $C_l$ is simply equal to its CPT $\mathcal{P}(C_l) = \mathbb{P}(R_l|S_l)$. The root node $C_r$ has no parent, so $\mathcal{P}(C_r) = \mathbb{P}(\ldots,(\oslash_{d_{j,r}}X_j),\ldots), \forall X_j \in C_r$. Furthermore, since $d_{j,i}$ is simply a count of how many times a variable in $C_i$ appears in each of the incident separators, the size of each tensor does not depend on which clique node was selected as the root.

**Example.** In Figure 2, node $C_{BCDE}$ corresponds to CPT $\mathbb{P}(B,D|C,E)$. Its high order tensor representation is $\mathcal{P}(C_{BCDE}) = \mathbb{P}(\oslash_2 B, D|\oslash_2 C, E)$, since both $B$ and $C$ occur twice in the separator sets incident to $C_{BCDE}$. Therefore the tensor $\mathcal{P}(\{B,C,D,E\})$ is a 6th order tensor with mode labels $\{B,B,D,C,C,E\}$.

### 4.2 Tensor message passing

With the higher order tensor representation for clique potentials in the junction tree as in (10), we can express the message update step in (8) as tensor multiplications. Consequently, we can compute the marginal distribution of the observed variables $\mathscr{O}$ in equation (7) recursively using a sequence of tensor multiplications. More specifically the general message update equation

for a node in a junction tree can be expressed as

$$\mathcal{M}(S_i) = \mathcal{P}(C_i) \times_{S_{i_1}} \mathcal{M}(S_{i_1}) \times_{S_{i_2}} \mathcal{M}(S_{i_2}). \quad (11)$$

Here the modes of the tensor $\mathcal{P}(C_i)$ are labeled by the variables, and the mode labels are used to carry out tensor multiplications as explained in Section 2. Essentially, multiplication with respect to the duplicated modes of the tensor $\mathcal{P}(C_i)$ will implement some kind of element-wise multiplication for the incoming messages and then summation over the variables in the remainder set $R_i$.

The tensor message passing steps in leaf nodes and the root node are special cases of the tensor message update in equation (11). The outgoing message $\mathcal{M}(S_l)$ at a leaf node $C_l$ can be computed by simply setting all variables in $R_l$ to the actual observed values $r_l$, i.e.,

$$\mathcal{M}(S_l) = \mathcal{P}(C_l)_{R_l = r_l} = \mathbb{P}(R_l = r_l | S_l). \quad (12)$$

In this step, there is no difference between the augmented tensor representation and the standard message passing in junction tree. At the root, we arrive at the final results of the message passing algorithm, and we obtain the marginal probability of the observed variables by aggregating all incoming messages from its 3 children, i.e.,

$$\mathbb{P}(\mathcal{O}) = \quad (13)$$
$$\mathcal{P}(C_r) \times_{S_{r_1}} \mathcal{M}(S_{r_1}) \times_{S_{r_2}} \mathcal{M}(S_{r_2}) \times_{S_{r_3}} \mathcal{M}(S_{r_3}).$$

**Example.** For Figure 2, using the following tensors

$$\mathcal{P}(\{B, C, D, E\}) = \mathbb{P}(\oslash_2 B, D \mid \oslash_2 C, E)$$
$$\mathcal{M}(\{B, C\}) = \mathbb{P}(f | B, C)$$
$$\mathcal{M}(\{B, D\}) = \mathbb{P}(g | B, D),$$

we can write the message update for node $C_{BCDE}$ in the form of equation (11) as

$$\mathcal{M}(\{C, E\}) = \mathcal{P}(\{B, C, D, E\}) \times_{\{B, C\}} \mathcal{M}(\{B, C\})$$
$$\times_{\{B, D\}} \mathcal{M}(\{B, D\}).$$

Note how the tensor multiplication sums out $B$ and $D$: $\mathcal{P}(\{B, C, D, E\})$ has two $B$ labels, and it appears in the subscripts of tensor multiplication twice; $D$ appears once in the label and in the subscript of tensor multiplication respectively. Similarly, $C$ is not summed out since there are two $C$ labels but it appears only once in the subscript of tensor multiplication.

## 5 Transformed Representation

Explicitly learning the tensor representation in (10) is still an intractable problem. Our key observation is that we do not need to recover the tensor representation explicitly if our focus is to perform inference using the message passing algorithm as in (11)–(13). As long as we can recover the tensor representation up to some invertible transformation, we can still obtain the correct marginal probability $\mathbb{P}(\mathcal{O})$.

More specifically, we can insert a mode-specific identity tensor $\mathcal{I}_\sigma$ into the message update equation in (11) without changing the outgoing message. Subsequently, we can then replace the mode-specific identity tensor by a pair of tensors, $\mathcal{F}$ and $\mathcal{F}^{-1}$, which are mode-specific inversions of each other ($\mathcal{F} \times_\omega \mathcal{F}^{-1} \cong \mathcal{I}_\sigma$). Then we can group these inserted tensors with the representation $\mathcal{P}(C)$ from (10), and obtain a transformed version $\widetilde{\mathcal{P}}(C)$ (also see Figure 1). Furthermore, we have the freedom in choosing these collections of tensor inversion pairs. We will show that if we choose them systematically, we will be able to estimate each transformed tensor $\widetilde{\mathcal{P}}(C)$ using the marginal probability of a small set of observed variables (observable representation). In this section, we will first explain the transformed tensor representation.

As an illustration, consider a sequence of matrix multiplications with two identity matrices $I_1 = F_1 F_1^{-1}$ and $I_2 = F_2 F_2^{-1}$ inserted

$$ABC = A(F_1 F_1^{-1}) B (F_2 F_2^{-1}) C$$
$$= \underbrace{(A F_1)}_{\widetilde{A}} \underbrace{(F_1^{-1} B F_2)}_{\widetilde{B}} \underbrace{(F_2^{-1} C)}_{\widetilde{C}}.$$

We see that we can equivalently compute $ABC$ using their transformed versions, i.e., $ABC = \widetilde{A}\widetilde{B}\widetilde{C}$.

Moving to the tensor case, let us first consider a node $C_i$ and its parent node $C_{i_0}$. Then the outgoing message of $C_{i_0}$ can be computed recursively as

$$\mathcal{M}(S_{i_0}) = \mathcal{P}(C_{i_0}) \times_{S_i} \underbrace{\mathcal{M}(S_i)}_{\mathcal{P}(C_i) \times_{S_{i_1}} \mathcal{M}(S_{i_1}) \times_{S_{i_2}} \mathcal{M}(S_{i_2})} \times \ldots$$

Inserting a mode specific identity tensor $\mathcal{I}_{S_i}$ with labels $\{S_i, S_i\}$ and similarly defined mode specific identity tensors $\mathcal{I}_{S_{i_1}}$ and $\mathcal{I}_{S_{i_2}}$ into the above two message updates, we obtain

$$\mathcal{M}(S_{i_0}) = \mathcal{P}(C_{i_0}) \times_{S_i} (\mathcal{I}_{S_i} \times_{S_i} \underbrace{\mathcal{M}(S_i)}_{\mathcal{P}(C_i) \times_{S_{i_1}} (\mathcal{I}_{S_{i_1}} \times_{S_{i_1}} \mathcal{M}(S_{i_1})) \times_{S_{i_2}} (\mathcal{I}_{S_{i_2}} \times_{S_{i_2}} \mathcal{M}(S_{i_1}))}) \times \ldots$$

Then we can further expand $\mathcal{I}_{S_i}$ using tensor inversion pairs $\mathcal{F}_i, \mathcal{F}_i^{-1}$, i.e., $\mathcal{I}_{S_i} = \mathcal{F}_i \times_{\omega_i} \mathcal{F}_i^{-1}$. Note that both $\mathcal{F}$ and $\mathcal{F}^{-1}$ have two set of mode labels, $S_i$ and another set $\omega_i$ which is related to the observable representation and explained in the next section. Similarly, we expand $\mathcal{I}_{S_{i_1}}$ and $\mathcal{I}_{S_{i_2}}$ using their corresponding tensor inversion pairs.

After expanding tensor identities $\mathcal{I}$, we can regroup terms, and at node $C_i$ we have

$$\mathcal{M}(S_i) = (\mathcal{P}(C_i) \times_{S_{i_1}} \mathcal{F}_{i_1} \times_{S_{i_2}} \mathcal{F}_{i_2}) \quad (14)$$
$$\times_{\omega_{i_1}} (\mathcal{F}_{i_1}^{-1} \times_{S_{i_1}} \mathcal{M}(S_{i_1}))$$
$$\times_{\omega_{i_2}} (\mathcal{F}_{i_2}^{-1} \times_{S_{i_2}} \mathcal{M}(S_{i_2}))$$

and at the parent node $C_{i_0}$ of $C_i$

$$\mathcal{M}(S_{i_0}) = (\mathcal{P}(C_{i_0}) \times_{S_i} \mathcal{F}_i \times \ldots) \quad (15)$$
$$\times_{\omega_i} (\mathcal{F}_i^{-1} \times_{S_i} \mathcal{M}(S_i)) \times \ldots$$

Now we can define the transformed tensor representation for $\mathcal{P}(C_i)$ as

$$\widetilde{\mathcal{P}}(C_i) := \mathcal{P}(C_i) \times_{S_{i_1}} \mathcal{F}_{i_1} \times_{S_{i_2}} \mathcal{F}_{i_2} \times_{S_i} \mathcal{F}_i^{-1}, \quad (16)$$

where the two transformations $\mathcal{F}_{i_1}$ and $\mathcal{F}_{i_2}$ are ob-

tained from the children side and the transformation $\mathcal{F}_i^{-1}$ is obtained from the parent side. Similarly, we can define the transformed representation for a leaf node and for the root node as

$$\widetilde{\mathcal{P}}(C_l) = \mathcal{P}(C_l) \times_{S_l} \mathcal{F}_l^{-1} \qquad (17)$$

$$\widetilde{\mathcal{P}}(C_r) = \mathcal{P}(C_r) \times_{S_{r_1}} \mathcal{F}_{r_1} \times_{S_{r_2}} \mathcal{F}_{r_2} \times_{S_{r_3}} \mathcal{F}_{r_3} \qquad (18)$$

Applying these definitions of the transformed representation recursively, we can perform message passing based purely on these transformed representations

$$\widetilde{\mathcal{M}}(S_{i_0}) = \widetilde{\mathcal{P}}(C_{i_0}) \times_{\boldsymbol{\omega}_i} \underbrace{\widetilde{\mathcal{M}}(S_i)}_{\widetilde{\mathcal{P}}(C_i) \times_{\boldsymbol{\omega}_{i_1}} \widetilde{\mathcal{M}}(S_{i_1}) \times_{\boldsymbol{\omega}_{i_2}} \widetilde{\mathcal{M}}(S_{i_2})} \times \ldots \qquad (19)$$

## 6 Observable Representation

In the transformed tensor representation in (16)-(18), we have the freedom of choosing the collection of tensor pairs $\mathcal{F}$ and $\mathcal{F}^{-1}$. We will show that if we choose them systematically, we can recover each transformed tensor $\widetilde{\mathcal{P}}(C)$ using the marginal probability of a small set of observed variables (observable representation).

We will focus on the transformed tensor representation in (16) for an internal node $C_i$ (other cases follow as special cases). Due to the recursive way the transformed representation is defined, we only have the freedom of choosing $\mathcal{F}_{i_1}$ and $\mathcal{F}_{i_2}$ in this formula; the choice of $\mathcal{F}_i$ will be fixed by the parent node of $C_i$. The idea is to choose

- $\mathcal{F}_{i_1} = \mathbb{P}(\mathscr{O}_{i_1}|S_{i_1})$ as the conditional distribution of some set of observed variables $\mathscr{O}_{i_1} \subset \mathscr{O}$ in the subtree rooted at child node $C_{i_1}$ of node $C_i$, conditioning on the corresponding separator set $S_{i_1}$.
- Similarly, we choose $\mathcal{F}_{i_2} = \mathbb{P}(\mathscr{O}_{i_2}|S_{i_2})$ where $\mathscr{O}_{i_2} \subset \mathscr{O}$ and it lies in subtree rooted at $C_{i_2}$.
- Following this convention, $\mathcal{F}_i$ is chosen by the parent node $C_{i_0}$ and is fixed to $\mathbb{P}(\mathscr{O}_i|S_i)$.

Therefore, we have

$$\widetilde{\mathcal{P}}(C_i) = \mathcal{P}(C_i) \times_{S_{i_1}} \mathbb{P}(\mathscr{O}_{i_1}|S_{i_1}) \times_{S_{i_2}} \mathbb{P}(\mathscr{O}_{i_2}|S_{i_2})$$
$$\times_{S_i} \mathbb{P}(\mathscr{O}_i|S_i)^{-1}, \qquad (20)$$

where the first two tensor multiplications essentially eliminate the latent variables in $S_{i_1}$ and $S_{i_2}$.[3] With these choices, we also fix the mode labels $\boldsymbol{\omega}_i$, $\boldsymbol{\omega}_{i_1}$ and $\boldsymbol{\omega}_{i_2}$ in (14) (15) and (19). That is $\boldsymbol{\omega}_i = \mathscr{O}_i$, $\boldsymbol{\omega}_{i_1} = \mathscr{O}_{i_1}$ and $\boldsymbol{\omega}_{i_2} = \mathscr{O}_{i_2}$.

To remove all dependencies on latent variables in $\widetilde{\mathcal{P}}(C_i)$ and relate it to observed variables, we need to eliminate the latent variables in $S_i$ and the tensor $\mathbb{P}(\mathscr{O}_i|S_i)^{-1}$. For this, we multiply the transformed tensor $\widetilde{\mathcal{P}}(C_i)$ by $\mathbb{P}(\mathscr{O}_i, \mathscr{O}_{i-})$, where $\mathscr{O}_{i-}$ denotes some set of observed variables which do *not* belong to the subtree rooted at node $C_i$. Furthermore, $\mathbb{P}(\mathscr{O}_i, \mathscr{O}_{i-})$ can be re-expressed using the conditional distribution

---
[3]If a latent variable in $S_{i_1} \cup S_{i_2}$ is also in $S_i$, it is not eliminated in this step but in another step.

---

of $\mathscr{O}_i$ and $\mathscr{O}_{i-}$ respectively, conditioning on the separator set $S_i$, i.e.,

$$\mathbb{P}(\mathscr{O}_i, \mathscr{O}_{i-}) = \mathbb{P}(\mathscr{O}_i|S_i) \times_{S_i} \mathbb{P}(\oslash_2 S_i) \times_{S_i} \mathbb{P}(\mathscr{O}_{i-}|S_i).$$

Therefore, we have

$$\mathbb{P}(\mathscr{O}_i|S_i)^{-1} \times_{\mathscr{O}_i} \mathbb{P}(\mathscr{O}_i, \mathscr{O}_{i-}) = \mathbb{P}(\oslash_2 S_i) \times_{S_i} \mathbb{P}(\mathscr{O}_{i-}|S_i),$$

and plugging this into (20), we have

$$\widetilde{\mathcal{P}}(C_i) \times_{\mathscr{O}_i} \mathbb{P}(\mathscr{O}_i, \mathscr{O}_{i-})$$
$$= \mathcal{P}(C_i) \times_{S_{i_1}} \mathbb{P}(\mathscr{O}_{i_1}|S_{i_1}) \times_{S_{i_2}} \mathbb{P}(\mathscr{O}_{i_2}|S_{i_2})$$
$$\quad \times_{S_i} \mathbb{P}(\oslash_2 S_i) \times_{S_i} \mathbb{P}(\mathscr{O}_{i-}|S_i)$$
$$= \mathbb{P}(\mathscr{O}_{i_1}, \mathscr{O}_{i_2}, \mathscr{O}_{i-}), \qquad (21)$$

where $\widetilde{\mathcal{P}}(C_i)$ is now related to only marginal probabilities of observed variables. From the equivalent relation, we can inverting $\mathbb{P}(\mathscr{O}_i, \mathscr{O}_{i-})$, and obtain the observable representation for $\widetilde{\mathcal{P}}(C_i)$

$$\widetilde{\mathcal{P}}(C_i) = \mathbb{P}(\mathscr{O}_{i_1}, \mathscr{O}_{i_2}, \mathscr{O}_{i-}) \times_{\mathscr{O}_{i-}} \mathbb{P}(\mathscr{O}_i, \mathscr{O}_{i-})^{-1}. \qquad (22)$$

**Example.** For node $C_{BCDE}$ in Figure 2, the choices of $\mathscr{O}_i, \mathscr{O}_{i_1}, \mathscr{O}_{i_2}$ and $\mathscr{O}_{i-}$ are $\{F, G\}$, $G$, $F$ and $H$ respectively.

There are many valid choices of $\mathscr{O}_{i-}$. In the supplementary, we describe how these different choices can be combined via a linear system using Eq. 21. This can substantially increase performance.

For the leaf nodes and the root node, the derivation for their observable representations can be viewed as special cases of that for the internal nodes. We provide the results for their observable representation below:

$$\widetilde{\mathcal{P}}(C_r) = \mathbb{P}(\mathscr{O}_{r_1}, \mathscr{O}_{r_2}, \mathscr{O}_{r_3}), \qquad (23)$$
$$\widetilde{\mathcal{P}}(C_l) = \mathbb{P}(\mathscr{O}_l, \mathscr{O}_{l-}) \times_{\mathscr{O}_{l-}} \mathbb{P}(\mathscr{O}_l, \mathscr{O}_{l-})^{-1}. \qquad (24)$$

If $\mathbb{P}(\mathscr{O}_l, \mathscr{O}_{l-})$ is invertible, then $\widetilde{\mathcal{P}}(C_l) = \mathcal{I}_{\mathscr{O}_l}$. Otherwise we need to project $\mathbb{P}(\mathscr{O}_i, \mathscr{O}_{i-})$ using a tensor $\mathcal{U}_i$ to make it invertible, as discussed in the next section. The overall algorithm is given in Algorithm 1. Given $N$ i.i.d. samples of the observed nodes, we simply replace $\mathbb{P}(\cdot)$ by the empirical estimate $\widehat{\mathbb{P}}(\cdot)$.

---

**Algorithm 1** Spectral algorithm for latent junction tree

**In**: Junction tree topology and $N$ i.i.d. samples $\{x_1^s, \ldots, x_{|\mathscr{O}|}^s\}_{s=1}^N$

**Out**: Estimated marginal $\widehat{\mathbb{P}}(\mathscr{O})$

1: Estimate $\widehat{\mathcal{P}}(C_i)$ for the root, leaf and internal nodes

$$\widehat{\mathcal{P}}(C_r) = \widehat{\mathbb{P}}(\mathscr{O}_{r_1}, \mathscr{O}_{r_2}, \mathscr{O}_{r_3}) \times_{\mathscr{O}_{r_1}} \mathcal{U}_{r_1} \times_{\mathscr{O}_{r_2}} \mathcal{U}_{r_2} \times_{\mathscr{O}_{r_3}} \mathcal{U}_{r_3}$$
$$\widehat{\mathcal{P}}(C_l) = \widehat{\mathbb{P}}(\mathscr{O}_l, \mathscr{O}_{l-}) \times_{\mathscr{O}_{l-}} (\widehat{\mathbb{P}}(\mathscr{O}_l, \mathscr{O}_{l-}) \times_{\mathscr{O}_l} \mathcal{U}_l)^{-1}$$
$$\widehat{\mathcal{P}}(C_i) = \widehat{\mathbb{P}}(\mathscr{O}_{i_1}, \mathscr{O}_{i_2}, \mathscr{O}_{i-}) \times_{\mathscr{O}_{i_1}} \mathcal{U}_{i_1} \times_{\mathscr{O}_{i_2}} \mathcal{U}_{i_2}$$
$$\quad \times_{\mathscr{O}_{i-}} (\widehat{\mathbb{P}}(\mathscr{O}_i, \mathscr{O}_{i-}) \times_{\mathscr{O}_i} \mathcal{U}_i)^{-1}$$

2: In reverse topological order, leaf and internal nodes send messages

$$\widehat{\mathcal{M}}(S_l) = \widehat{\mathcal{P}}(C_l)_{\mathscr{O}_l = o_l}$$
$$\widehat{\mathcal{M}}(S_i) = \widehat{\mathcal{P}}(C_i) \times_{\mathscr{O}_{i_1}} \widehat{\mathcal{M}}(S_{i_1}) \times_{\mathscr{O}_{i_2}} \widehat{\mathcal{M}}(S_{i_2})$$

3: At the root, obtain $\widehat{\mathbb{P}}(\mathscr{O})$ by

$$\widehat{\mathcal{P}}(C_r) \times_{\mathscr{O}_{r_1}} \widehat{\mathcal{M}}(S_{r_1}) \times_{\mathscr{O}_{r_2}} \widehat{\mathcal{M}}(S_{r_2}) \times_{\mathscr{O}_{r_3}} \widehat{\mathcal{M}}(S_{r_3})$$

## 7 Discussion

The observable representation exists only if there exist tensor inversion pairs $\mathcal{F}_i = \mathbb{P}(\mathcal{O}_i|S_i)$, and $\mathcal{F}_i^{-1}$. This is equivalent to requiring that the rank of the matricized version of $\mathcal{F}_i$ (rows corresponds to modes $\mathcal{O}_i$ and column to modes $S_i$) has rank $\tau_i := k_h \times |S_i|$. Similarly, the matricized version of $\mathbb{P}(\mathcal{O}_{-i}|S_i)$ also needs to have rank $\tau_i$, so that the matricized version of $\mathbb{P}(\mathcal{O}_i, \mathcal{O}_{i-})$ has rank $\tau_i$ and is invertible. Thus, it is required that #states($\mathcal{O}_i$) ≥ #states($S_i$). This can be achieved by either making $\mathcal{O}_i$ consist of a few high dimensional observations, or of many smaller dimensional ones. In the case when #states($\mathcal{O}_i$) > #states($S_i$), we need to project $\mathcal{F}_i$ to a lower dimensional space using a tensor $\mathcal{U}_i$ so that it can be inverted. In this case, we define $\mathcal{F}_i := \mathbb{P}(\mathcal{O}_i|S_i) \times_{\mathcal{O}_i} \mathcal{U}_i$. For example, following this through the computation for the leaf gives us that $\widetilde{\mathcal{P}}(C_l) = \mathbb{P}(\mathcal{O}_l, \mathcal{O}_{l-}) \times_{\mathcal{O}_{l-}} (\mathbb{P}(\mathcal{O}_l, \mathcal{O}_{l-}) \times_{\mathcal{O}_l} \mathcal{U}_l)^{-1}$. A good choice of $\mathcal{U}_i$ can be obtained by performing a singular value decomposition of the matricized version of $\mathbb{P}(\mathcal{O}_i, \mathcal{O}_{i-})$ (variables in $\mathcal{O}_i$ are arranged to rows and those in $\mathcal{O}_{i-}$ to columns).

For HMMs and latent trees, this rank condition can be expressed simply as requiring the conditional probability tables of the underlying model to not be rank-deficient. However, junction trees encode more complex latent structures that introduce subtle considerations. A general characterization of the existence condition for observable representation with respect to the graph topology will be our future work. In the appendix, we give some intuition using a couple of examples where observable representations do not exist.

## 8 Sample Complexity

We analyze the sample complexity of Algorithm 1 and show that it depends on the junction tree topology and the spectral properties of the true model. Let $d_i$ be the order of $\mathcal{P}(C_i)$ and $e_i$ be the number of modes of $\mathcal{P}(C_i)$ that correspond to observed variables.

**Theorem 1** Let $\tau_i = k_h \times |S_i|$, $d_{\max} = \max_i d_i$, and $e_{\max} = \max_i e_i$. Then, for any $\epsilon > 0, 0 < \delta < 1$, if

$$N \geq O\left(\left(\frac{4k_h^2}{3\beta^2}\right)^{d_{\max}} \frac{k_o^{e_{\max}} \ln \frac{|\mathbb{C}|}{\delta}|\mathbb{C}|^2}{\epsilon^2 \alpha^4}\right)$$

where $\sigma_\tau(*)$ returns the $\tau^{th}$ largest singular value and

$$\alpha = \min_i \ \sigma_{\tau_i}(\mathbb{P}(\mathcal{O}_i, \mathcal{O}_{-i})), \quad \beta = \min_i \ \sigma_{\tau_i}(\mathcal{F}_i)$$

Then with probability $1 - \delta$,
$\sum_{x_1,\ldots,x_{|\mathcal{O}|}} \left|\widehat{\mathbb{P}}(x_1,\ldots,x_{|\mathcal{O}|}) - \mathbb{P}(x_1,\ldots,x_{|\mathcal{O}|})\right| \leq \epsilon$ .

See the supplementary for a proof. The result implies that the estimation problem depends exponentially on $d_{max}$ and $e_{max}$, but note that $e_{max} \leq d_{max}$. Furthermore, $d_{max}$ is always greater than or equal to the treewidth. Note the dependence on the singular values of certain probability tensors. In fully observed models, the accuracy of the learned parameters depends only on how close the empirical estimates of the factors are to the true factors. However, our spectral algorithm also depends on how close the inverses of these empirical estimates are to the true inverses, which depends on the spectral properties of the matrices (Stewart & Sun, 1990).

## 9 Experiments

We now evaluate our method on synthetic and real data and compare it with both standard EM (Dempster et al., 1977) and stepwise online EM (Liang & Klein, 2009). All methods were implemented in C++, and the matrix library Eigen (Guennebaud et al., 2010) was used for computing SVDs and solving linear systems. For all experiments, standard EM is given 5 random restarts. Online EM tends to be sensitive to the learning rate, so it is given one restart for each of 5 choices of the learning rate $\{0.6, 0.7, 0.8, 0.9, 1\}$ (the one with highest likelihood is selected). Convergence is determined by measuring the change in the log likelihood at iteration $t$ (denoted by $f(t)$) over the average: $\frac{|f(t)-f(t-1)|}{\text{avg}(f(t),f(t-1))} \leq 10^{-4}$ (the same precision as used in Murphy (2005)).

For large sample sizes our method is almost two orders of magnitude faster than both EM and online EM. This is because EM is iterative and every iteration requires inference over all the training examples which can become expensive. On the other hand, the computational cost of our method is dominated by the SVD/linear system. Thus, it is primarily dependent only on the number of observed states and maximum tensor order, and can easily scale to larger sample sizes.

In terms of accuracy, we generally observe 3 distinct regions, low-sample size, mid-sample size, and large sample size. In the low sample size region, EM/online EM tend to overfit to the training data and our spectral algorithm usually performs better. In the mid-sample size region EM/online EM tend to perform better since they benefit from a smaller number of parameters. However, once a certain sample size is reached (the large sample size region), our spectral algorithm consistently outperforms EM/online EM which suffer from local minima and convergence issues.

### 9.1 Synthetic Evaluation

We first perform a synthetic evaluation. 4 different latent structures are used (see Figure 3): a second order nonhomogenous (NH) HMM, a third order NH HMM, a 2 level NH factorial HMM, and a complicated synthetic junction tree. The second/third order HMMs have $k_h = 2$ and $k_o = 4$, while the factorial HMM and synthetic junction tree have $k_h = 2$, and $k_o = 16$. For each latent structure, we generate 10 sets of model parameters, and then sample $N$ training points and 1000 test points from each set, where $N$ is varied from 100 to 100,000. For evaluation, we measure the accuracy of joint estimation using $error = \frac{|\widehat{\mathbb{P}}(x_1,...,x_O) - \mathbb{P}(x_1,...,x_O)|}{\mathbb{P}(x_1,...,x_O)}$. We also measure the training time of both methods.

Figure 3 shows the results. As discussed earlier, our algorithm is between one and two orders of magnitude faster than both EM and online EM for all the latent structures. EM is actually slower for very small sample sizes than for mid-range sample sizes because of overfitting. Also, in all cases, the spectral algorithm has the lowest error for large sample sizes. Moreover, critical sample size at which spectral overtakes EM/online EM is largely dependent on the number of parameters in the observable representation compared to that in the original parameterization of the model. In higher order/factorial HMM models, this increase is small, while in the synthetic junction tree it is larger.

### 9.2 Splice dataset

We next consider the task of determining splicing sites in DNA sequences (Asuncion & Newman, 2007). Each example consists of a DNA sequence of length 60, where each position in the sequence is either an $A$, $T$, $C$, or $G$. The goal is to classify whether the sequence is an Intron/Exon site, Exon/Intron site, or neither. During training, for each class a different second order nonhomogeneous HMM with $k_h = 2$ and $k_o = 4$ is trained. At test, the probability of the test sequence is computed for each model, and the one with the highest probability is selected (which we found to perform better than a homogeneous one).

Figure 4, shows our results, which are consistent with our synthetic evaluation. Spectral performs the best in low sample sizes, while EM/online EM perform a little better in the mid-sample size range. The dataset is not large enough to explore the large sample size regime. Moreover, we note that spectral algorithm is much faster for all the sample sizes.

## 10 Conclusion

We have developed an alternative parameterization that allows fast, local minima free, and consistent parameter learning of latent junction trees. Our approach generalizes spectral algorithms to a much wider range of structures such as higher order, factorial, and semi-hidden Markov models. Unlike traditional nonconvex optimization formulations, spectral algorithms allow us to theoretically explore latent variable models in more depth. The spectral algorithm depends not only on the junction tree topology but also on the spectral properties of the parameters. Thus, two models with the same structure may pose different degrees of difficulty based on the underlying singular values. This is very different from learning fully observed junction trees, which is primarily dependent on only the topol-ogy/treewidth. Future directions include learning discriminative models and structure learning.

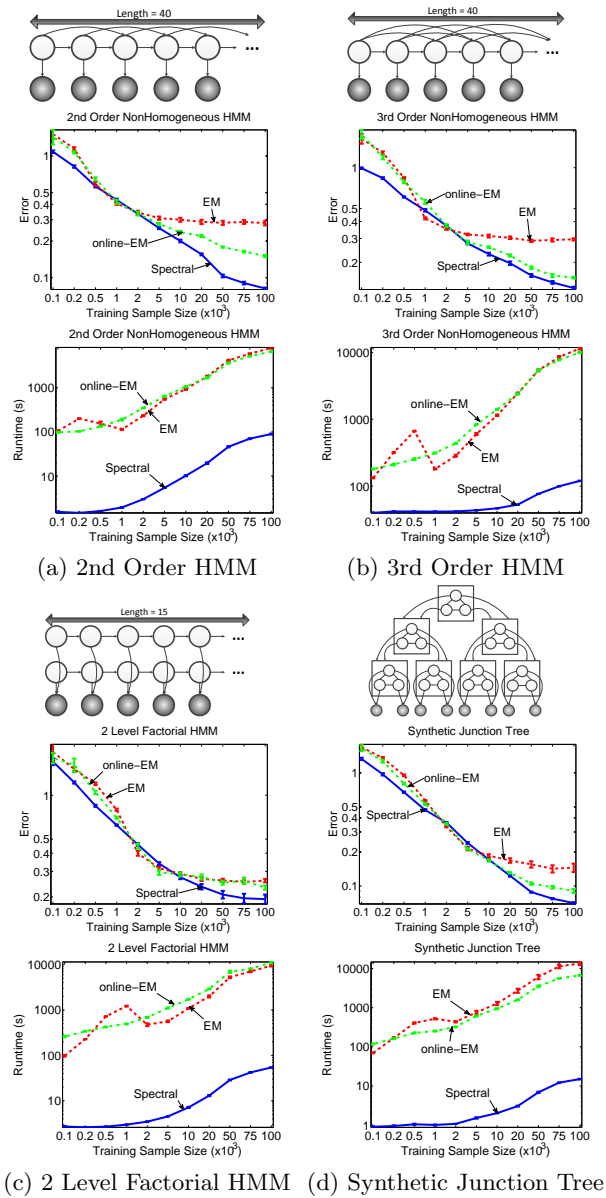

(a) 2nd Order HMM  (b) 3rd Order HMM

(c) 2 Level Factorial HMM  (d) Synthetic Junction Tree

Figure 3: Comparison of our spectral algorithm (blue) to EM (red) and online EM (green) for various latent structures. Both errors and runtimes in log scale.

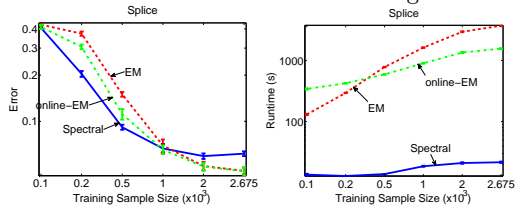

Figure 4: Results on Splice dataset

**Acknowledgements:** This work is supported by an NSF Graduate Fellowship (Grant No. 0750271) to APP, Georgia Tech Startup Funding to LS, NIH 1R01GM093156, and The Gatsby Charitable Foundation. We thank Byron Boots for valuable discussion.

# References

Asuncion, A. and Newman, D.J. UCI machine learning repository, 2007.

Balle, B., Quattoni, A., and Carreras, X. A spectral learning algorithm for finite state transducers. *Machine Learning and Knowledge Discovery in Databases*, pp. 156–171, 2011.

Blei, David and McAuliffe, Jon. Supervised topic models. In *Advances in Neural Information Processing Systems 20*, pp. 121–128. 2007.

Blei, D.M., Ng, A.Y., and Jordan, M.I. Latent dirichlet allocation. *The Journal of Machine Learning Research*, 3:993–1022, 2003.

Cohen, S.B., Stratos, K., Collins, M., Foster, D.P., and Ungar, L. Spectral learning of latent-variable pcfgs. In *Association of Computational Linguistics (ACL)*, volume 50, 2012.

Cowell, R., Dawid, A., Lauritzen, S., and Spiegelhalter, D. *Probabilistic Networks and Expert Sytems*. Springer, New York, 1999.

Dempster, A., Laird, N., and Rubin, D. Maximum likelihood from incomplete data via the EM algorithm. *Journal of the Royal Statistical Society B*, 39(1):1–22, 1977.

Ghahramani, Z. and Jordan, M.I. Factorial hidden Markov models. *Machine learning*, 29(2):245–273, 1997.

Guennebaud, G., Jacob, B., et al. Eigen v3. http://eigen.tuxfamily.org, 2010.

Hsu, D., Kakade, S., and Zhang, T. A spectral algorithm for learning hidden Markov models. In *Proc. Annual Conf. Computational Learning Theory*, 2009.

Kolda, T. and Bader, B. Tensor decompositions and applications. *SIAM Review*, 51(3):455–500, 2009.

Koller, D. and Friedman, N. *Probabilistic graphical models: principles and techniques*. The MIT Press, 2009.

Kundu, A., He, Y., and Bahl, P. Recognition of handwritten word: first and second order hidden Markov model based approach. *Pattern recognition*, 22(3):283–297, 1989.

Lacoste-Julien, S., Sha, F., and Jordan, M.I. Disclda: Discriminative learning for dimensionality reduction and classification. volume 21, pp. 897–904. 2008.

Liang, P. and Klein, D. Online em for unsupervised models. In *Proceedings of human language technologies: The 2009 annual conference of the North American chapter of the association for computational linguistics*, pp. 611–619. Association for Computational Linguistics, 2009.

Murphy, K. Hidden Markov model (HMM) toolbox for matlab http://www.cs.ubc.ca/murphyk/software/. 2005.

Murphy, K.P. *Dynamic bayesian networks: representation, inference and learning*. PhD thesis, University of California, 2002.

Parikh, A.P., Song, L., and Xing, E.P. A spectral algorithm for latent tree graphical models. In *Proceedings of the 28th International Conference on Machine Learning*, pp. 1065–1072. ACM, 2011.

Rabiner, L. R. and Juang, B. H. An introduction to hidden Markov models. *IEEE ASSP Magazine*, 3(1):4–16, 1986.

Song, L., Boots, B., Siddiqi, S., Gordon, G., and Smola, A. Hilbert space embeddings of hidden Markov models. In *Proceedings of the 27th International Conference on Machine Learning*, pp. 991–998. ACM, 2010.

Song, L., Parikh, A.P., and Xing, E.P. Kernel embeddings of latent tree graphical models. In *Advances in Neural Information Processing Systems (NIPS)*, volume 24, pp. 2708–2716. 2011.

Stewart, GW and Sun, J. *Matrix Perturbation Theory*. Academic Press, 1990.

Su, X. and Khoshgoftaar, T.M. A survey of collaborative filtering techniques. *Advances in Artificial Intelligence*, 2009:4, 2009.

Zhu, J., Ahmed, A., and Xing, E.P. Medlda: maximum margin supervised topic models for regression and classification. In *Proceedings of the 26th Annual International Conference on Machine Learning*, pp. 1257–1264. ACM, 2009.